\documentclass{article}

\usepackage{microtype}
\usepackage{graphicx,amsmath}
\usepackage{amssymb}
\usepackage{calc}
\usepackage[normalem]{ulem}
\usepackage{mathtools} 
\usepackage{enumitem}

\usepackage{amsthm}
\newtheorem{theorem}{Theorem}

\usepackage{booktabs} % for professional tables

\usepackage{hyperref}

\usepackage[accepted]{icml2019}

\icmltitlerunning{On the Convex Behavior of Deep Neural Networks in Relation to the Layers' Width}
\begin{document}

\twocolumn[
\icmltitle{On the Convex Behavior of Deep Neural Networks \\
           in Relation to the Layers' Width}



\icmlsetsymbol{equal}{*}

\begin{icmlauthorlist}
\icmlauthor{Etai Littwin}{tau}
\icmlauthor{Lior Wolf}{tau,fb}
\end{icmlauthorlist}

\icmlaffiliation{tau}{Tel Aviv University}
\icmlaffiliation{fb}{Facebook AI Research}

\icmlkeywords{Machine Learning, ICML}

\vskip 0.3in
]



\printAffiliationsAndNotice{\icmlEqualContribution} % otherwise use the standard text.

\begin{abstract}

The Hessian of neural networks can be decomposed into a sum of two matrices: (i) the positive semidefinite generalized Gauss-Newton matrix $G$, and (ii) the matrix $H$ containing negative eigenvalues. We observe that for wider networks, minimizing the loss with the gradient descent optimization maneuvers through surfaces of positive curvatures at the start and end of training, and close to zero curvatures in between. In other words, it seems that during crucial parts of the training process, the Hessian in wide networks is dominated by the component $G$.

To explain this phenomenon, we show that when initialized using common methodologies, the gradients of over-parameterized networks are approximately orthogonal to $H$, such that the curvature of the loss surface is strictly positive in the direction of the gradient. 
\end{abstract}

\section{Introduction}
It has frequently been observed~\cite{chrom} that deep neural networks are relatively easy to optimize using straightforward SGD algorithms, despite having complex non-convex loss surfaces. A great body of work in recent years has focused on the properties of SGD and its ability to maneuver through saddle points and complex surface topologies, in order to explain the relative ease of the optimization process~\cite{sgd}. A complementary approach to analyzing the optimization process, is to consider the trajectories of the gradients during optimization, with the guiding intuition that while the loss surface defined by modern neural networks might be incredibly complex and hard to analyze, the trajectories of gradient descent on these loss surfaces might prove rather simple. 

An additional direction of research focuses on the spectrum of the Hessian of the loss during training, as the knowledge of the second order information about the loss surface can tell us quite a bit about the behavior of the optimizer, and how we could modify our algorithms to make them converge faster and to better solutions~\cite{shampoo, fisher}. Recent works have empirically shown that the distribution of the eigenvalues of the Hessian is composed of a bulk, which is concentrated around zero, and the edges which are scattered away from zero~\cite{hess}. Modern neural networks, however, contain millions of parameters, resulting in a hefty Hessian with millions of eigenvalues, the overwhelming majority of which are irrelevant. 

In our work, we seek to study the curvature of the Hessian in directions corresponding to the trajectory of the gradients during training, by focusing on the Hessian gradient product. Specifically, we aim to empirically and theoretically analyze the path of gradient descent algorithms by analyzing the Hessian at initialization.

We identify and empirically demonstrate novel phenomena relating to the trajectory of gradient descent neural networks under certain conditions.  We observe that the gradients follow a convex path during training when performing SGD, such that the curvature of the loss is seldom negative in the direction of the gradients. This effect is more prominent in wider networks, i.e., in CNNs with a larger number of kernels. %For SGD, we show that the gradients on a mini-batch is in a direction of positive curvature of its corresponding loss. 
Furthermore, we theoretically prove that when using common initialization methods, the gradients of random feedforward networks are indeed pointing in a direction of positive curvature when the network is wide enough, and the input is large enough. 

\subsection{Terminology}
Given the output of some neural network $y(x_i,w_t)$ for input $x_i$, with $W_t$ denoting the model weights at iteration $t$, and a convex, twice differentiable loss function $\mathcal{L}(y(x_i,W_t))$, we denote the gradient on a batch of $N$ samples by $g_t = \frac{1}{N}\sum_{i=1}^N \frac{\partial \mathcal{L}(y(x_i,W_t))}{\partial W_t}$, and the Hessian by $\mathcal{H}_t = \frac{1}{N}\sum_{i=1}^N \frac{\partial^2 \mathcal{L}(y(x_i,W_t))}{\partial W_t \partial W_t}$. A positive curvature during training at step $t$ is characterized by the following condition:
\begin{equation}
    g_t^\top \mathcal{H}_tg_t \geq 0
\end{equation}

\begin{figure*}[t]
%\vskip 0.2in
\begin{center}
\begin{tabular}{c@{}c}
\includegraphics[trim=30 20 30 30, clip,width=1.051\columnwidth]{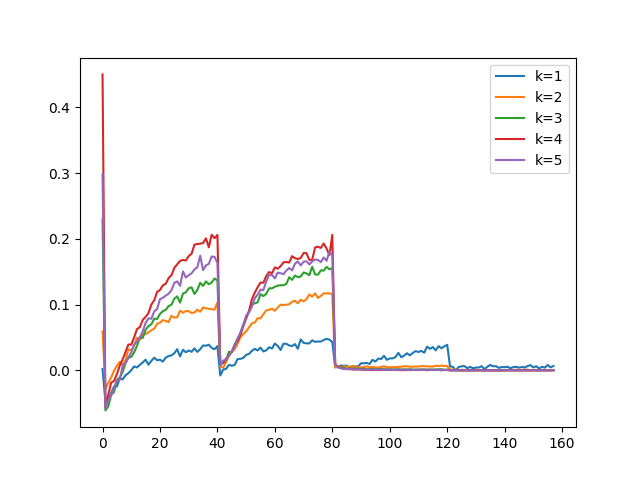}&
\includegraphics[trim=30 20 30 30, clip,width=1.051\columnwidth]{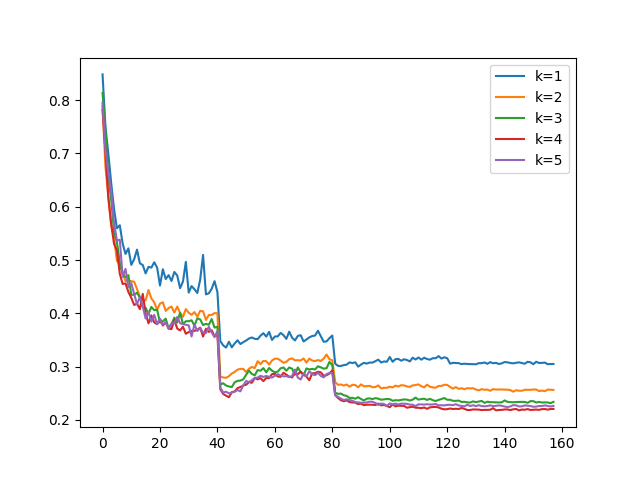}\\
(a) & (b)\\
\includegraphics[trim=30 20 30 30, clip,width=1.051\columnwidth]{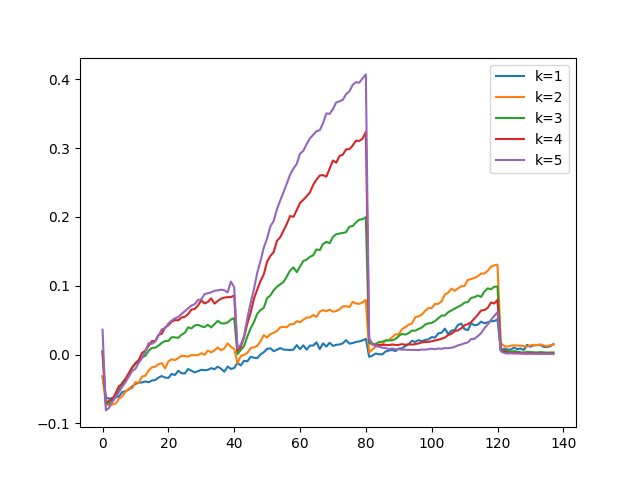}&
\includegraphics[trim=30 20 30 30, clip,width=1.051\columnwidth]{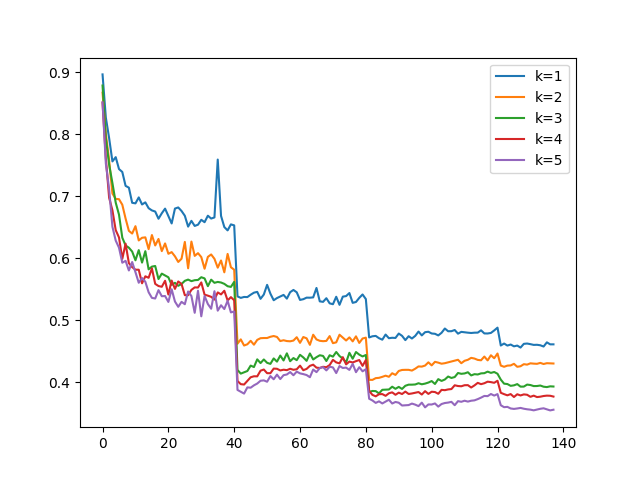}\\
(c) & (d)\\
\end{tabular}
\caption{The approximated curvature in the direction of the gradient, for networks with different width multiplier $K=1..5$.  (a,c) as a function of the epoch for multiple network widths, shown for CIFAR-100, and tiny imagenet, respectively. The different plots denote different layer width. (b,d) the test loss for the same networks trained on CIFAR-100 and tiny imagenet.}
\label{fig:main}
\end{center}
\vskip -0.2in
\end{figure*}

\begin{figure*}[t]
\vskip 0.1in
\begin{center}
\begin{tabular}{cc}
\includegraphics[trim=33 20 30 30, clip,width=.4994932\linewidth]{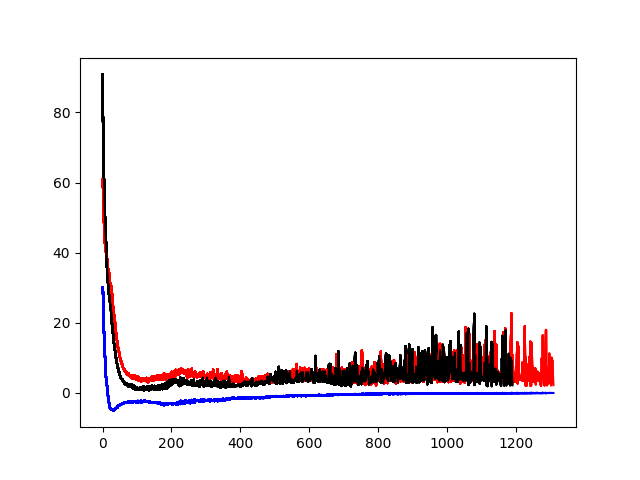} &
\includegraphics[trim=33 20 30 30, clip,width=.4994932\linewidth]{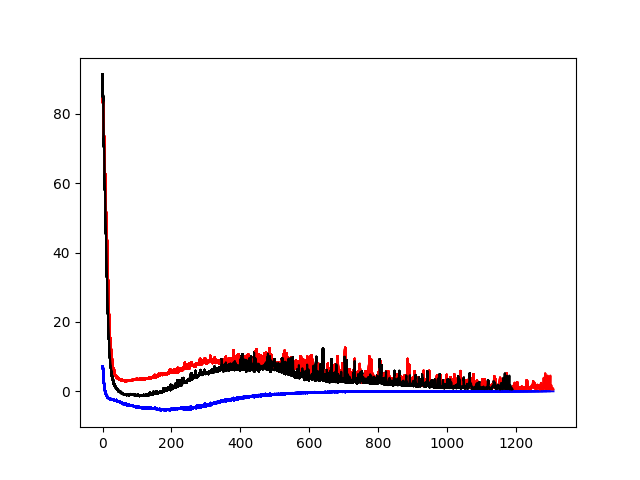}\\
(a) & (b)\\
\includegraphics[trim=33 20 30 30, clip,width=.4994932\linewidth]{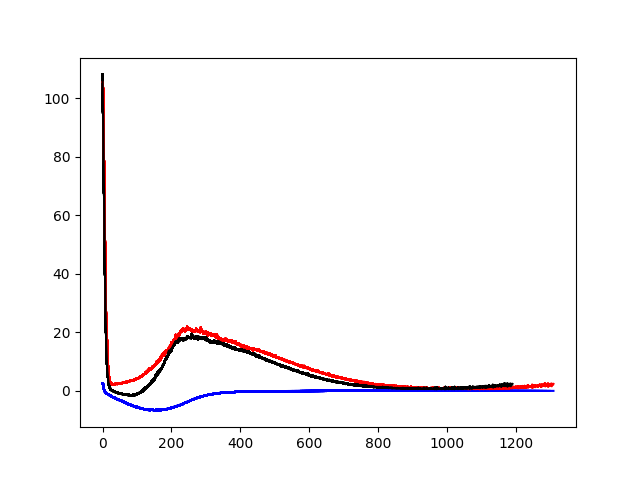}\\
(c)\\
%\includegraphics[width=\columnwidth]{figures/train_400.png}\\
%(a) & (b) & (c)\\
\end{tabular}
\caption{The curvature in the direction of the gradient, decomposed to its components $\mathcal{H}_{\hat{g}}$(black), $G_{\hat{g}}$(red) and $H_{\hat{g}}$(blue), as a function of epoch,  for fully connected networks with layers' width of (a) 50, (b) 200, and (c) 400, trained on random data and labels. In all cases, $\mathcal{H}_{\hat{g}}$ is mostly dominated by $G_{\hat{g}}$, with $H_{\hat{g}}$ starting closer to zero for wider networks.}
\label{fig:toy}
\end{center}
\vskip -0.2in
\end{figure*}

\section{Observations}
We approximate the Hessian gradient product $g_t^\top \mathcal{H}_tg_t$ during the training of deep neural networks, using the following formula:
\begin{multline}
    \mathcal{L}(W_{t+1},\{x_i\}_{i=1}^N) \approx \mathcal{L}(W_t,\{x_i\}_{i=1}^N)\\ - \delta\|g_t\|^2 + \delta^2g_t^\top \mathcal{H}_tg_t
    \end{multline}
    
    Which directly leads to: \begin{multline}
    g_t^\top \mathcal{H}_tg_t  \approx \frac{\mathcal{L}(W_{t+1},\{x_i\}_{i=1}^N)-\mathcal{L}(W_t,\{x_i\}_{i=1}^N)  }{\delta^2}\\
    +\frac{\|g_t\|^2}{\delta}
\end{multline}
where $\delta$ is the learning rate, and the loss $\mathcal{L}$ is computed on a minibatch $\{x_i\}_{i=1}^N$. 

We track the behavior of $g_t^\top \mathcal{H}_tg_t$ across the epochs for networks of varying widths. In our experiments, we employ the ResNet-50 network~\cite{resnet}, which has layers with 16, 32, or 64 kernels. We make these layers wider by multiplying the number of kernels by a factor $K=1,\dots,5$. We halve the learning rate at epochs 40, 80, and 120. A batch size of 100 is used with a vanilla SGD optimizer. 

The results are shown in Fig.~\ref{fig:main} for two popular datasets: CIFAR-100~\cite{cifar} and tiny imagenet~\cite{tiny}. In both datasets, one can observe that (i) throughout the majority of the training run, $g_t^\top \mathcal{H}_tg_t$ remains positive; (ii) at the start of training, and when the network converges, it obtains larger values; and (iii) typically, the wider the network, the bigger this quantity is. 

Fig.~\ref{fig:main} also presents the test error for the same benchmarks and the same networks. We observe that wider networks lead to smaller test errors, which means that their tendency toward a convex-mimicking behavior during training does not come at the expanse of the generalization ability.%up until convergence, with a slight dip into negative values early in training. We also observe that the wider the network, the less time it spends in "negative curvature" territory.

We further conduct small scale experiments, in which one can efficiently decompose the empirical Hessian into its components $H$ and $G$, and calculate the quantities $H_{\hat{g}} = \hat{g}^\top H\hat{g}$, $G_{\hat{g}} = \hat{g}^\top G \hat{g}$ and $\mathcal{H}_{\hat{g}} = \hat{g}^\top \mathcal{H} \hat{g}$ during training. Specifically, we train a single output, 8 layer fully connected network with layer width of 50, 200 and 400 on a set of 1000 predetermined random inputs sampled from a Gaussian distribution with dimensionality matching the width of the network, which are then normalized to have unit norm. The network are trained to fit with a standard L2 loss random labels sampled from a binary distribution. 

The results are shown in Fig.~\ref{fig:toy}. As can be seen, in all cases, the network exhibits a convex-mimicking behaviour, similar to the results of our CNN experiments. Specifically, the curvature observed in $\mathcal{H}_{\hat{g}}$ seems to closely follow that of $G_{\hat{g}}$. In addition, as predicted in our analysis, the curve of $H_{\hat{g}}$ starts closer to zero for wider networks. 

\section{Analysis}

An initial analysis is performed for linear networks (no non-linearities) at initialization, where the weights of the networks are randomly sampled. We make very few assumptions on the nature of the networks' initialization, and the popular initialization techniques satisfy our assumptions. %{\color{red}An extension to non-linear networks is discussed at the end of our analysis.}

\subsection{Overview}
In our theoretical analysis, we focus on the Hessian of a single output {\color{black}linear network} with $L$ layers, where the output of layer $l=1,\dots,L$ is given by $y^l = W^{l\top}y^{l-1}$ where $W^l \in \mathbb{R}^{n_{l-1}\times n_l}, n_L = 1, y^l \in \mathbb{R}^{n_l}$, and the input, which is also denoted by $y^0$ is given by a fixed dataset $\{x_s\}_{s=1}^N, x_s \in \mathbb{R}^{n_0}$. We denote by $W$ the concatenation of all the parameters of the network in a column vector. 

As mentioned, we consider a twice differentiable convex loss function $\mathcal{L}$. We further assume that the loss has bounded first and second derivatives. The Hessian matrix is given by:
\begin{multline}
    \mathcal{H} = \sum_{s=1}^N \mathcal{L}''(y(x_s)^L)g_w^s g_w^{s\top} + \sum_{s=1}^N \mathcal{L}'(y(x_s)^L)\frac{\partial ^2 y^L(x_s)}{\partial W \partial W}\\
    = \sum_{s=1}^N \mathcal{L}''(y(x_s)^L)\hat{G}^s + \sum_{s=1}^N \mathcal{L}'(y(x_s)^L)\hat{H}^s
     = G + H
\end{multline}

where $g_w^s$ is the gradient of the network output unit for input $s$ with respect to the weights. The first term of the Hessian is comprised of a sum of positive semidefinite matrices, and is, therefore, positive semidefinite, while the second term is not, and is dependent on the architecture. In the following, we aim to show that under certain conditions pretaining to the width of the network and the dimensionality  of the input, which are usually met in practice, the trajectory of GD is influenced primarily by $G$. That is, denoting by index $t$ the evaluation of a tensor at iteration $t$, it holds that:
\begin{equation}
    g_t^\top \mathcal{H}_t g_t \approx g_t^\top G_t g_t > 0
\end{equation}
\subsection{Main Results}
In our analysis, we consider the weights $W^k = [w_{ij}^k]_{ 0<i\leq n_k,0<j\leq n_{k-1},0<k\leq L}$ sampled iid from a symmetric distribution with moments $m_t$, such that $m_2,m_4,m_6 \approx \frac{1}{n_{k-1}}$, where $m_t = \mathbb{E}\left( (w_{ij}^k)^t \right)$ are the moments of the distribution. Notice that this condition applies to the common initialization scheme~\cite{init}. We first consider the gradient and Hessian evaluated on a single sample with unit norm, i.e., $\|y^0\| = 1$, and then discuss multiple samples. We make the following notations: the derivative of the output unit with respect to a single weight $ij$ in layer $k$ is denoted by $g_{ij}^k = \frac{\partial y^L}{\partial w_{ij}^k} = \frac{\partial y^L}{\partial y^k}\frac{\partial y^k}{\partial w_{ij}^k}$.

  we denote $\hat{H}$ by:
  \begin{equation}
  \hat{H}_{ijk}^{uvl} = \frac{\partial^2 y^L}{\partial w_{ij}^k \partial w_{uv}^l} 
\end{equation}
where the triplets of indices $ijk,uvl$ parameterize row and column indices in the matrix $\hat{H}$ (note that the ranges of $ij,uv$ depend on the layer indices $k$ and $l$).
The next theorem gives the expected mean and variance of $g^\top \hat{H} g$ for wide networks at initialization:

\begin{theorem}\label{the:1}
$\mathbb{E}\left(g^T \hat{H} g\right) = 0$ and for $min_{l<L}(n_l)>>L$, it holds that
$ \mathbb{E}\left((g^T \hat{H} g)^2\right) \approx \Theta(\frac{\sum_{k=1}^L\sum_{k'=1}^L n_kn_{k'}}{n_0^3})$
\end{theorem}

The Hessian gradient product is not necessarily indicative of the local curvature, since we do not take into account the norm of the gradient. As shown in~\cite{over}, a constant width network ensures that the norm of the gradient approximately equals the input. More specifically, by having $\forall_{0 \leq l \leq L},~ n_l = nm_l$, we ensure that $\|g\|^2 \rightarrow \frac{\sum_{l=1}^L m_l}{m_0}\|y^0\|^2$ as $n$ gets larger. (The precise statement is omitted for brevity). The next theorem gives a bound on the probability of the local curvature to be above some $\epsilon$, given the full Hessian.
\begin{theorem}
Assume that $\forall_{0 \leq l \leq L},~ n_l = nm_l$, and $Prob(\frac{\sum_{l=1}^L m_l}{m_0}\|y^0\|^2 - \epsilon \leq \|g\|^2 \leq \frac{\sum_{l=1}^L m_l}{m_0}\|y^0\|^2 + \epsilon) >1 - \delta(\epsilon)$, $\mathcal{L}''(y^L)>\alpha>0,\mathcal{L}'(y^L)>\beta>0$  then it holds that:
\begin{multline}
Prob \bigg(\hat{g}^\top \mathcal{H} \hat{g} >\epsilon \bigg)
\geq (1-\delta(\frac{2\epsilon}{\|\alpha\|}))(1-\delta(\epsilon))\\ (1 - \frac{1}{n}\frac{\gamma\sum_{kk'=1}^L m_km_{k'}}{m_0^3(\frac{\epsilon}{\|\beta\|})^2(\frac{\sum_{l=1}^L m_l}{m_0}\|y^0\|^2 - (\frac{\epsilon}{\|\beta\|}))^2})
\end{multline}
\end{theorem}

The above analysis provides a rather surprising result, which is that the dimensionality of the input also plays a role in determining the local curvature at initialization, in addition to the width of the network. 

%{\color{red}Since the input is fixed, it seems we would have little control over the local curvature of the loss at the start of training for fully connected networks. The case, however, is different for CNNs. While the input is fixed, we do have control over the receptive field of the first layer, causing a similar effect as changing the input dimensionality. This behaviour is also observed in our CNN experiments, where we demonstrate that the first layer has a distinct roll in facilitating a fast convergence. A larger receptive field in the first layer causes a more rapid start to training, where a similar change to subsequent layers changes little.}

We have discussed a single sample in our analysis so far. In the case of linear neural networks, the result of Thm.~\ref{the:1} also hold for the batch case since we did not assume anything on the input. Specifically, since $H = \sum_{s=1}^N \mathcal{L}'(y(x_s)^L)\hat{H}^s$, and considering the gradient of sample $u$, and $\hat{H}^v$ evaluated on sample $v$, then it holds that $g^{u\top}\hat{H}^v g^u \approx \frac{\sum_{kk'}n_kn_{k'}}{n_0^3}$. 

The term $g^\top G g$ can be significantly smaller in the batch case, even when the gradient is large. Indeed, under mild conditions, when gradients of individual samples are orthogonal to each other, this term can be made small while keeping the overall gradient large, insuring a faster convergence to the optimum. Precise statement and analysis are kept for future work.
 %This implies that lowering the batch size would lead to 

% Denoting the loss of the network with weights $W$ evaluated on some batch of $N$ examples $\{x_i\}_{i=1}^N$ at iteration $t$ by $\mathcal{L}_t(W_t,\{x_i\}_{i=1}^N)$, then for a small enough learning rate $\delta$, it holds at iteration $t+1$:
% \begin{multline}
%     \mathcal{L}_{t+1}(W_{t+1},\{x_i\}_{i=1}^N) \approx \mathcal{L}_t(W_t,\{x_i\}_{i=1}^N) - \delta\|g_t\|^2 + g_t^\top \mathcal{H}_tg_t\\ \rightarrow
%     g_t^\top \mathcal{H}_tg_t \approx \mathcal{L}_{t+1}(W_{t+1},\{x_i\}_{i=1}^N)-\mathcal{L}_t(W_t,\{x_i\}_{i=1}^N)+\delta\|g_t\|^2
% \end{multline}
% and so we can easily evaluate the local curvature in the direction of the gradient by checking the difference between the decrease in loss, and the norm of the gradient multiplied by the learning rate.

\section{Discussion}

The literature has already linked the width of a neural network to the property of norm-preserving. For example,~\cite{over} show that under different sets of assumptions, the wider the network, the more likely the norm of the activations will be preserved with depth. Our work is the first one, as far as we know, to show a link between a convex-mimicking behavior during optimization and the network's width.

The literature also teaches us about the properties of the Hessian during training~\cite{hess}. As far as we know, we are unique in that we look at the Hessian in the direction of the gradient, where it matters most to the optimization process.

The line of research can lead to a better understanding of the link between different architectures and their trainability. Moreover, the signal we track can be readily used in order to drive the optimization process. Specifically, it may be derivable to adjust the hyperparameters (learning rate, batch size, batch composition, etc.) such that the gradients are large while the product of the gradient with the Hessian implies close to zero curvature. 

\section*{Acknowledgements}
This project has received funding from the European Research Council (ERC) under the European
Union’s Horizon 2020 research and innovation programme (grant ERC CoG 725974).

%\clearpage
\bibliography{relu}
\bibliographystyle{icml2019}

\clearpage
\appendix
\onecolumn
\section{Proof of Thm.~1}

\begin{proof}
% The Hessian gradient product $\sum_{ijk}\hat{H}_{ijk}^{uvl}g_{ij}^k$ depends on $l$. For $l=1$, it is $\sum_i(a_{2i})^2 y_v^0y^1_j\delta_{ju} + \sum_{ijk>2}(a_{ki})^2a_{lu}^{k-1j}y_v^{0}y^{k-1}_j$; for $l=2$: $\sum_j a_{2u}a_{1i} (y_j^0)^2\delta_{vi} + \sum_i (a_{3i})^2 y_v^{1}y_j^2\delta_{ju} + \sum_{ijk>3}(a_{ki})^2a_{2u}^{k-1j}y_v^1y^{k-1}_j$; in the case of $L-1>l>2$, it is given by $\sum_{ijk>l+1}(a_{ki})^2a_{lu}^{k-1j}y_v^{l-1}y^{k-1}_j + \sum_{ijk<l-1}a_{lu} a_{ki}^{l-1v}a_{ki}(y_j^{k-1})^2 + \sum_i(a_{l+1i})^2y_v^{l-1}y_j^l\delta_{ju} + \sum_ja_{lu}a_{l-1i}(y_j^{l-2})^2\delta_{vi}$; for the case of $l=L-1$, $\sum_j (a_{Lj})^2 y_v^{L-2}y_j^{L-1}\delta_{ju} + \sum_{ij} a_{L-1u}a_{L-2i} (y_j^{L-3})^2\delta_{vi}\\ + \sum_{ijk<L-2}a_{L-1u}a_{ki} a_{ki}^{L-2v}(y_j^{k-1})^2$; and, finally, for the case of $l=L$: $\sum_{ij} a_{Lu} a_{L-1i} (y_j^{L-2})^2\delta_{vi} + \sum_{ijk<L-1}a_{Lu}a_{ki} a_{ki}^{L-1v}(y_j^{k-1})^2$. 

 The derivative of unit $j$ in layer $k$ with respect to unit $u$ in layer $l$ is denoted by $\frac{\partial y_j^k}{\partial y_u^l} = a_{lu}^{kj}$. Dropping the unit indexes $j$ or $u$ means deriving the entire layer, or with respect to the entire layer respectively. Dropping all superscripts means derivative of the output unit, so that $\frac{\partial y_L}{\partial y_u^l} = a_{lu}$. It then holds that $g_{ij}^k = a_{ki}y^{k-1}_j$. Finally, we denote by $W_{,i}^l$ and $W_{i,}^l$ the $i'th$ column and row vectors of $W^l$ respectively.
We have:
\begin{equation}
  \hat{H}_{ijk}^{uvl} = \frac{\partial^2 y^L}{\partial w_{ij}^k \partial w_{uv}^l} = \begin{array}{cc}
  \{ & 
    \begin{array}{cc}
      a_{ki} a_{lu}^{k-1j}y_v^{l-1} & l<k-1 \\
      a_{ki} y_v^{l-1}\delta_{ju} & l=k-1 \\
      0 & l=k \\
      a_{lu} y_j^{k-1}\delta_{vi} & l=k+1 \\
      a_{lu} a_{ki}^{l-1v}y_j^{k-1} & l>k+1 \\
    \end{array}
\end{array}
\end{equation}
where the triplets of indices $ijk,uvl$ parameterize row and column indices in the matrix $\hat{H}$ (note that the ranges of $ij,uv$ depend on the layer indices $k$ and $l$).

The Hessian gradient product is given by:
\begin{equation}
    \sum_{ijk}\hat{H}_{ijk}^{uvl}g_{ij}^k = \begin{array}{cc}
  \{ & 
    \begin{array}{cc}
      \sum_i(a_{2i})^2 y_v^0y^1_j\delta_{ju} + \sum_{ijk>2}(a_{ki})^2a_{lu}^{k-1j}y_v^{0}y^{k-1}_j & l=1 \\
      \sum_j a_{2u}a_{1i} (y_j^0)^2\delta_{vi} + \sum_i (a_{3i})^2 y_v^{1}y_j^2\delta_{ju} + \sum_{ijk>3}(a_{ki})^2a_{2u}^{k-1j}y_v^1y^{k-1}_j & l=2 \\
      \sum_{ijk>l+1}(a_{ki})^2a_{lu}^{k-1j}y_v^{l-1}y^{k-1}_j + \sum_{ijk<l-1}a_{lu} a_{ki}^{l-1v}a_{ki}(y_j^{k-1})^2 \\+ \sum_i(a_{l+1i})^2y_v^{l-1}y_j^l\delta_{ju} + \sum_ja_{lu}a_{l-1i}(y_j^{l-2})^2\delta_{vi} & L-1>l>2 \\
      \sum_j (a_{Lj})^2 y_v^{L-2}y_j^{L-1}\delta_{ju} + \sum_{ij} a_{L-1u}a_{L-2i} (y_j^{L-3})^2\delta_{vi}\\ + \sum_{ijk<L-2}a_{L-1u}a_{ki} a_{ki}^{L-2v}(y_j^{k-1})^2  & l=L-1 \\
      \sum_{ij} a_{Lu} a_{L-1i} (y_j^{L-2})^2\delta_{vi} + \sum_{ijk<L-1}a_{Lu}a_{ki} a_{ki}^{L-1v}(y_j^{k-1})^2 & l=L \\
    \end{array}
\end{array}
\end{equation}
The claim $\mathbb{E}\left(g^T \hat{H} g\right) = 0$ is trivially true, since each term in the Hessian gradient products will contain odd moments of the weights, which zero out for symmetric distributions. 
For deep networks, we have that:
\begin{equation}
\mathbb{E}\left((g^\top \hat{H} g)^2\right)  = \mathbb{E}\left((\sum_{uvl}\sum_{ijk}g_{uv}^l\hat{H}_{ijk}^{uvl}g_{ij}^k)^2\right) \approx   \mathbb{E}\left((\sum_{uv,L-1>l>2}\sum_{ij,k\neq [l-1,l+1]}g_{uv}^l\hat{H}_{ijk}^{uvl}g_{ij}^k)^2\right)
\end{equation}
and so:
\begin{multline}
    \sum_{\substack{uv,\\L-1>l>2}}\sum_{\substack{ij,\\k\neq [l-1,l+1]}}g_{uv}^l\hat{H}_{ijk}^{uvl}g_{ij}^k = \sum_{uvijk>l+1}(a_{ki})^2a_{lu}a_{lu}^{k-1j}(y_v^{l-1})^2y^{k-1}_j + \sum_{uvijk<l-1}(a_{lu})^2 a_{ki}^{l-1v}a_{ki}(y_j^{k-1})^2y_v^{l-1}
\end{multline}
\begin{multline}\label{eq:top}
    \mathbb{E}\left((g^\top \hat{H} g)^2\right)  \approx \mathbb{E}\left((\sum_{uvijk>l+1}(a_{ki})^2a_u^la_{lu}^{k-1j}(y_v^{l-1})^2y^{k-1}_j + \sum_{uvijk<l-1}(a_{lu})^2 a_{ki}^{l-1v}a_{ki}(y_j^{k-1})^2y_v^{l-1})^2 \right)\\
    \approx \mathbb{E}\left((\sum_{uvijk>l+1}\sum_{u'v'i'j'k'>l'+1}(a_{ki})^2(a_{k'i'})^2a_{lu}a_{l'u'}a_{lu}^{k-1j}a_{l'u'}^{k'-1j'}(y_v^{l-1})^2(y_{v'}^{l'-1})^2y^{k-1}_jy^{k'-1}_{j'}\right)\\
    \approx \mathbb{E}\left(\sum_{uvij u'v'i'j}(a_{ki})^2(a_{k'i'})^2a_{lu}a_{l'u'}a_{lu}^{k-1j}a_{l'u'}^{k'-1j'}(y_v^{l-1})^2(y_{v'}^{l'-1})^2y^{k-1}_jy^{k'-1}_{j'}\right)
\end{multline}
where we assumed in the last transition that $min_{l<L}(n_l)>>L$.
Applying expectation on the last weight layer $W^L$, it holds that:
\begin{multline}
    \mathbb{E}\left(\sum_{uvij u'v'i'j}(a_{ki})^2(a_{k'i'})^2a_{lu}a_{l'u'}\right) \\ \approx \frac{1}{n_{L-1}^3} \mathbb{E}\bigg( \sum_{uvij u'v'i'j}(\|a_{ki}^{L-1}\|^2\|a_{k'i'}^{L-1}\|^2a_{lu}^{L-1\top}a_{l'u'}^{L-1}
    + (a_{ki}^{L-1\top} a_{k'i'}^{L-1})^2a_{lu}^{L-1\top}a_{l'u'}^{L'-1}\\
    + (a_{ki}^{L-1\top} a_{k'i'}^{L-1})a_{k'i'}^{L'-1\top}a_{lu}^{L-1} a_{ki}^{L-1\top}a_{l'u'}^{L'-1} + (a_{ki}^{L-1\top} a_{k'i'}^{L-1})a_{k'i'}^{L'-1\top}a_{l'u'}^{L'-1} a_{ki}^{L-1\top}a_{lu}^{L-1}) \bigg)
\end{multline}  
Plugging each term in Eq.~\ref{eq:top} yields the same asymptotic behaviour. For the sake of brevity, we will show this with the first term. Without loss of generality, we assume $k>k'+1>l+2>l'+3$, we now recursively apply expectations over $W^{L-1}...W^{k}$, while noticing that the following hold for fixed vectors $v_1...v_6$:
\begin{multline}
  \mathbb{E}\bigg(v_1^\top W^{l\top} W^l v_2 v_3^\top W^{l\top} W^l v_4 v_5^\top W^{l\top} W^l v_6\bigg) \approx \frac{n_l^3}{n_{l-1}^3}v_1^\top v_2 v_3^\top  v_4 v_5^\top  v_6\\
  \mathbb{E}\bigg(v_1^\top W_{j,}^{l\top} W_{,j}^l v_2 v_3^\top W^{l\top} W^l v_4 v_5^\top W^{l\top} W^l v_6\bigg) \approx \frac{n_l^2}{n_{l-1}^3}v_1^\top v_2 v_3^\top  v_4 v_5^\top  v_6\\
  \mathbb{E}\bigg(W_{,j}^{l\top} W^l v_1 W_{,j}^{l\top} W^l v_2 v_3^\top W^{l\top} W^l v_4\bigg) \approx \frac{n_l^2}{n_{l-1}^3}v_1^\top v_2 v_3^\top  v_4 
\end{multline}
and so:
\begin{multline}
    \mathbb{E}\left((g^\top \hat{H} g)^2\right)
    \approx  \frac{1}{n_{k}^3} \mathbb{E}\bigg( \sum_{uvij u'v'i'j}(\|a_{ki}^{k}\|^2\|a_{k'i'}^{k}\|^2a_{lu}^{k\top}a_{l'u'}^{k}a_{lu}^{k-1j}a_{l'u'}^{k'-1j'}(y_v^{l-1})^2(y_{v'}^{l'-1})^2y^{k-1}_jy^{k'-1}_{j'})\bigg)\\
     \approx  \frac{1}{n_{k-1}^3} \mathbb{E}\bigg( \sum_{uvij u'v'i'j}(\|a_{k'i'}^{k-1}\|^2a_{lu}^{k-1\top}a_{l'u'}^{k-1}a_{lu}^{k-1j}y^{k-1}_ja_{l'u'}^{k'-1j'}(y_v^{l-1})^2(y_{v'}^{l'-1})^2y^{k'-1}_{j'})\bigg)\\
     \approx \frac{1}{n_{k-1}n_{k-2}^3} \mathbb{E}\bigg( \sum_{uvij u'v'i'j}(\|a_{k'i'}^{k-2}\|^2a_{lu}^{k-2\top}a_{l'u'}^{k-2}a_{lu}^{k-2\top}y^{k-2} a_{l'u'}^{k'-1j'}(y_v^{l-1})^2(y_{v'}^{l'-1})^2y^{k'-1}_{j'})\bigg)
\end{multline}
Recursively going through $l=k-2...1$ we have:
\begin{equation}
    \mathbb{E}\left((g^\top \hat{H} g)^2\right)\approx \frac{\|y^0\|^6}{n_{k-1}n_{k'-1}n_{l-1}n_{l'-1}n_l n_{l'}n_0^3}\sum_{uvij u'v'i'j} = \frac{n_{k-1}n_kn_{k'-1}n_k'n_{l-1}n_ln_{l'-1}n_l'}{n_{k-1}n_{k'-1}n_{l-1}n_{l'-1}n_l n_{l'}n_0^3}\\
     = \frac{n_kn_{k'}}{n_0^3}
\end{equation}
Summing over $k,k'$:
\begin{equation}
\mathbb{E}\left((g^\top \hat{H} g)^2\right)\approx \frac{\sum_{kk'}n_kn_{k'}}{n_0^3}
\end{equation}
\end{proof}

\section{Proof of Thm.~2}
\begin{proof}
We have:
\begin{multline}
    Prob \bigg(\hat{g}^\top \mathcal{H} \hat{g} >\epsilon \bigg) \geq Prob \bigg(\hat{g}^\top G \hat{g} >2\epsilon \bigg)Prob \bigg(|\hat{g}^\top H \hat{g}|<\epsilon \bigg)\\
     = Prob \bigg(\mathcal{L}''\|g\|^2 >2\epsilon \bigg)Prob \bigg(|\mathcal{L}'||\hat{g}^\top \hat{H} \hat{g}| <\epsilon \bigg) \geq Prob \bigg(|\alpha|\|g\|^2 >2\epsilon \bigg)Prob \bigg(|\beta||\hat{g}^\top \hat{H} \hat{g}| <\epsilon \bigg)
\end{multline}
\begin{multline}
    Prob \bigg(|\hat{g}^\top \hat{H} \hat{g}| < \epsilon \bigg)\\ \geq   Prob \bigg(\frac{1}{\frac{\sum_{l=1}^L m_l}{m_0}\|y^0\|^2 + \epsilon} \leq \frac{1}{\|g\|^2} \leq     \frac{1}{\frac{\sum_{l=1}^L m_l}{m_0}\|y^0\|^2 - \epsilon} \bigg)Prob \bigg(|g^\top \hat{H} g| < \epsilon(\frac{\sum_{l=1}^L m_l}{m_0}\|y^0\|^2 - \epsilon) \bigg)
\end{multline}

From The.~\ref{the:1}, we have that $var(g^\top \hat{H} g) = \frac{\gamma\sum_{kk'=1}^L m_km_{k'}}{nm_0^3}$ for some $\gamma>0$. Using Chebyshev's inequality:
\begin{equation}
    Prob \bigg(|g^\top \hat{H} g| < \epsilon(\frac{\sum_{l=1}^L m_l}{m_0}\|y^0\|^2 - \epsilon) \bigg) > 1 - \frac{1}{n}\frac{\gamma\sum_{kk'=1}^L m_km_{k'}}{m_0^3\epsilon^2(\frac{\sum_{l=1}^L m_l}{m_0}\|y^0\|^2 - \epsilon)^2}
\end{equation}
and so:
\begin{equation}
    Prob \bigg(|\hat{g}^\top \hat{H} \hat{g}| < \epsilon \bigg) \geq (1-\delta(\epsilon))(1 - \frac{1}{n}\frac{\gamma\sum_{kk'=1}^L m_km_{k'}}{m_0^3\epsilon^2(\frac{\sum_{l=1}^L m_l}{m_0}\|y^0\|^2 - \epsilon)^2})
\end{equation}
Finally:
\begin{equation}
    Prob \bigg(\hat{g}^\top \mathcal{H} \hat{g} >\epsilon \bigg) \geq (1-\delta(\frac{2\epsilon}{\|\alpha\|}))(1-\delta(\epsilon))(1 - \frac{1}{n}\frac{\gamma\sum_{kk'=1}^L m_km_{k'}}{m_0^3(\frac{\epsilon}{\|\beta\|})^2(\frac{\sum_{l=1}^L m_l}{m_0}\|y^0\|^2 - (\frac{\epsilon}{\|\beta\|}))^2})
\end{equation}
\end{proof}

\end{document}